\documentclass[11pt,a4paper]{article}
\usepackage{times,latexsym}
\usepackage{url}
\usepackage[T1]{fontenc}
\usepackage[acceptedWithA]{tacl2021v1}
\hypersetup{hypertexnames=false}
\usepackage{graphicx}
\usepackage{booktabs}
\usepackage{array}
\usepackage{amsmath}
\usepackage{float}
\usepackage{placeins}

\def\ie{\emph{i.e.}, }

\makeatletter
\long\def\@makefntext#1{%
  \parindent 0pt%
  \noindent
  \ifx\@thefnmark\@empty
    #1%
  \else
    \textsuperscript{\normalfont\@thefnmark}\,#1%
  \fi
}
\makeatother
\newcommand{\unmarkedfootnotetext}[1]{%
  \begingroup
  \renewcommand{\thefootnote}{}%
  \footnotetext{#1}%
  \endgroup
}

\makeatletter
\newcommand{\arxivtitlefigure}{%
  \vskip 0.18in
  \begin{minipage}{\textwidth}
    \centering
    \includegraphics[width=0.98\textwidth]{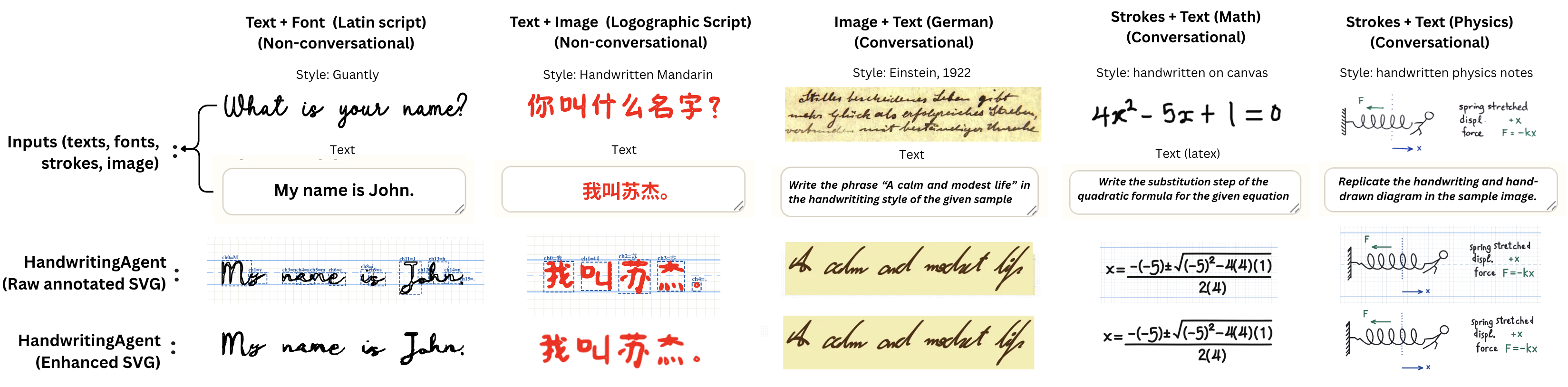}%
    \def\@captype{figure}%
    \caption{HandwritingAgent synthesizes natural, style-consistent handwriting sequences by reasoning over the glyph structure, stroke geometry, and writing dynamics of a target handwriting style in natural language. Given a reference handwriting style in image format and conditioned on texts (in either conversational and non-conversational modes), HandwritingAgent can synthesize diverse handwriting styles across multiple scripts and tasks, including Latin, logographic, and symbolic expressions, without the need for task-specific training.}
    \label{fig:overview}
  \end{minipage}
}
\makeatother

\newcommand{\ablationstyleimage}[1]{%
  \raisebox{-0.18\height}{\includegraphics[width=0.65in,height=0.18in,keepaspectratio]{#1}}%
}
\newcommand{\ablationresultimage}[1]{%
  \includegraphics[width=1.00in,height=0.25in,keepaspectratio]{#1}%
}

\newcommand{\figref}[1]{\hyperref[#1]{Figure~\ref*{#1}}}
\newcommand{\tabref}[1]{\hyperref[#1]{Table~\ref*{#1}}}
\newcommand{\secref}[1]{\hyperref[#1]{Section~\ref*{#1}}}
\newcommand{\tabtworef}[2]{\hyperref[#1]{Tables~\ref*{#1}} and \hyperref[#2]{\ref*{#2}}}

\title{HandwritingAgent: Language-Driven Handwriting Synthesis in\\Scalable Vector Space}
\author{Jaward Sesay$^{1}$\quad Yue Yu$^{1\dagger}$\quad B{\"o}rje F. Karlsson$^{2}$\\[0.45em]
  $^{1}$Beijing Institute of Technology\\
  $^{2}$Beijing Academy of Artificial Intelligence}
\date{}

\begin{document}
\maketitle
\begingroup
\renewcommand{\thefootnote}{\fnsymbol{footnote}}
\footnotetext[2]{Corresponding author.}
\endgroup
\unmarkedfootnotetext{\raggedright Open-source code repository for this work: \href{https://github.com/Jaykef/HandwritingAgent}{https://github.com/Jaykef/HandwritingAgent}.\par}

\begin{abstract}
Teaching machines to emulate natural handwriting styles remains an open challenge, as it requires synthesizing stroke sequences that dynamically vary in shape, texture, pressure and script—not only across individuals but also within a single person's handwriting. Attempts at this challenge have largely explored deep learning methods in both online and offline settings. However, these approaches are often constrained by style-specific architectural choices, heavy reliance on large datasets, high compute costs, and a lack of flexible control over writing styles through natural language. To this end, we introduce \textbf{HandwritingAgent}, a language-driven agent that can synthesize natural handwriting sequences directly in Scalable Vector Graphics (SVG) format with no need for style-specific training. The agent leverages a large reasoning model to geometrically analyse and autoregressively generate target handwritten glyphs as stroke sequences in a discrete grid canvas environment. Generation is conditioned on texts provided in either conversational or non-conversational mode, along with a reference handwriting-style image. Experiments on diverse handwriting tasks spanning imitation, recognition, multi-lingual handwriting synthesis, and generation of complex handwritten maths and science expressions indicate substantial improvement in performance, with HandwritingAgent matching or surpassing state-of-the-art generative handwriting models, while providing a more efficient, controllable, and generalizable synthesis method.
\end{abstract}

\section{Introduction}

Handwriting remains a symbolic form of expression in human language, shaped through education, culture, and neurophysiological processes that develop over time~\citep{vanGalen1991,Dinehart2015}. For centuries it has served as a primary medium for preserving knowledge, expressing thoughts and emotions, and the passing down of cultural values across generations~\citep{Fischer2001}. Studies in graphonomics show that one's handwriting emerges from complex motor programs that evolve through years of practice, resulting in a uniquely personalized writing style~\citep{Kao1986,vanGemmert2015}. These personal attributes make handwriting inherently difficult to model in machines, particularly in important practical domains such as identity verification, forgery detection, and recognition, where issues of authenticity, variability, and data scarcity remain persistent concerns~\citep{Elarian2014}.

Handwriting synthesis emerged as a technological response to these challenges, with early approaches relying on explicit structural, parametric, and rule-based representations to model character forms and stroke composition~\citep{Guyon1996}, while subsequent work demonstrated that writer-specific styles could be generated from limited samples by modelling variability through statistical and dynamic programming techniques~\citep{Chowdhury2009,Haines2016}. Related efforts extended focus to scripts with stronger morphological dependencies such as Arabic \citep{Elarian2015} and Chinese \citep{Lyu2017}.

The advent of deep learning shifted synthesis from hand-crafted modelling to data-driven generative methods that learn writing patterns directly from large corpora, RNN- and LSTM-based models were among the first to generate text-conditioned continuous handwriting trajectories~\citep{Graves2013}, followed by GANs that enabled diverse handwritten text image generation with controllable style and content~\citep{Alonso2019,Kang2020}, diffusion-based models producing high-quality samples through iterative denoising~\citep{Luhman2020}, and transformer-based architectures capturing long-range dependencies in styled handwriting generation~\citep{Bhunia2021}. Collectively, these methods yielded substantial improvements in visual realism, stylistic controllability, and generalization to arbitrary text and image modalities.

However, despite these advances, existing handwriting synthesis methods exhibit several limitations. For instance, dominant deep learning approaches depend on large training corpora and substantial compute, which makes adaptation to new writers, styles, or low-resource settings more difficult. Additionally, most architectures operate in raster space, producing outputs that are difficult to control at the stroke level because they lack explicit geometric data. Furthermore, most methods are specialized for particular scripts or languages, requiring retraining or architectural changes to support new writing scripts. These limitations highlight a broader challenge: existing handwriting synthesis pipelines rely primarily on data-driven pattern learning, rather than reasoning about the structural and geometric principles underlying handwriting in natural language.

\begin{figure}[H]
  \centering
  \includegraphics[width=1.0\columnwidth]{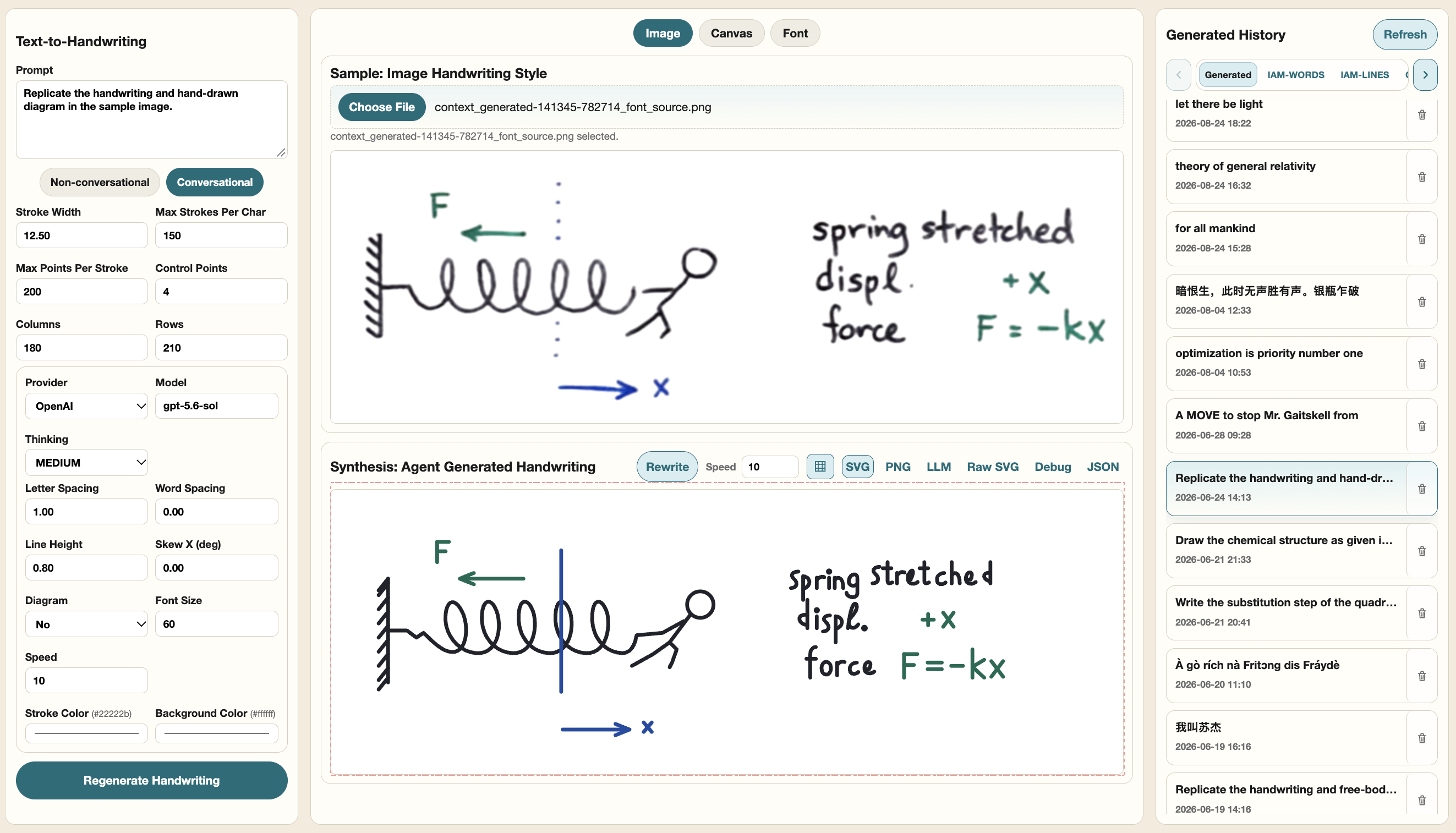}
  \caption{Interactive handwriting synthesis environment.}
  \label{fig:hwstudio}
\end{figure}

Recent advances in large language models (LLMs) have enabled strong performance in structured reasoning, symbolic manipulation, and controllable synthesis in natural language. When deployed as agents operating over explicit representations, these models can address tasks that require compositional reasoning and constraint satisfaction more efficiently than purely data-driven approaches. Building on this, we propose \textbf{HandwritingAgent}, a language-driven agent that discretely synthesizes handwriting sequences directly in SVG space without any style-specific training. As summarized in \figref{fig:overview}, given a target handwriting style in image or stroke form, the agent leverages an underlying language model to reason on geometric cues, plan glyph formation, and adapt target writing style to that of the sample. The desired handwriting is then generated sequentially at the stroke level in vectorized form within a discrete grid-canvas environment. This approach recasts handwriting synthesis as a reasoning-guided symbolic generation problem. As a result, the agent can adapt efficiently to new styles, generalize across scripts or other domains, and generate interpretable outputs that are useful for iterative refinement. Our main contributions are as follows:

\begin{enumerate}
  \item Training-free adaptation: The agent can adapt to new handwriting styles directly from a small set of samples, with no need for style-specific training.
  \item Multilingual and multi-domain generalization: It can generalize across multiple languages and writing tasks, spanning Latin and logographic scripts, as well as symbolic domains that involve generating handwritten maths and science expressions.
  \item Interpretable outputs: The agent synthesizes handwritten text discretely in SVG format, which is inherently resolution-independent, editable, and interpretable.
  \item Synthesis environment: An interactive handwriting synthesis environment (shown in \figref{fig:hwstudio}) that allows for easy use of the underlying framework providing user-friendly features for generating, analysing, and previewing diverse handwriting styles.
\end{enumerate}

\section{Related Work}

\subsection{Handwriting Synthesis Via Generative Modelling}

As neural networks became more capable of modelling complex sequential and visual patterns, efforts in handwriting synthesis increasingly adopted generative modelling architectures that learn the distribution of handwriting styles from data. Recent generative methods have largely explored transformer, GAN, and diffusion-based architectures. Handwriting Transformers (HWT) use self-attention to capture local and global style patterns for few-shot handwritten text generation~\citep{Bhunia2021}, while VATr++~\citep{Vanherle2024} builds on a hybrid convolutional-transformer architecture with improved input preparation and training regularization strategies to enhance rare character generalization across unseen styles. Emuru~\citep{Pippi2025} and Eruku~\citep{Zaccagnino2026} extend this line of work with autoregressive latent-image generation, improving zero-shot generalization to unseen styles, generation length flexibility, and text adherence. In parallel, diffusion-based methods generate handwriting through iterative denoising under style and content constraints. DiffusionPen~\citep{Nikolaidou2024} adopts a few-shot latent diffusion formulation with explicit style encoding for controllable writer-conditioned generation, and One-DM~\citep{Dai2024} further reduces style conditioning to a single reference sample while improving fine-grained style extraction and robustness to background noise. While these methods have significantly improved style adaptation and generation quality, they remain constrained by large-scale training data, compute, and limited control over style using language.

\subsection{Generating Scalable Vector Graphics}

Scalable Vector Graphics (SVG) is a structured XML-based vector graphics format for representing two-dimensional visual content through explicit geometric instructions that define paths, lines, curves, shapes, and text. Unlike raster images, SVG preserves the internal geometry of curves, strokes, and spatial composition. This makes it especially suitable for tasks in which the structure of the written trace matters as much as its appearance. Prior efforts have explored vector graphics generation as sequential prediction over drawing commands~\citep{Lopes2019}, and hierarchical models such as DeepSVG showed that complex vector compositions can be synthesized while maintaining structural coherence~\citep{Carlier2020}. Other studies generate vector graphics from raster inputs without direct vector supervision~\citep{Reddy2021}, while work in font modelling demonstrated that vector outlines can be synthesized consistently at the glyph level~\citep{Wang2021}. Recent language-guided SVG systems further suggest that vector space supports a more symbolic form of visual generation, one naturally compatible with editability, compositional control, and geometric reasoning \citep{Jain2022VectorFusion,Xing2024SVGDreamer}.

\subsection{Language-driven Handwriting Synthesis}

\begin{figure}[t]
  \centering
  \includegraphics[width=\columnwidth]{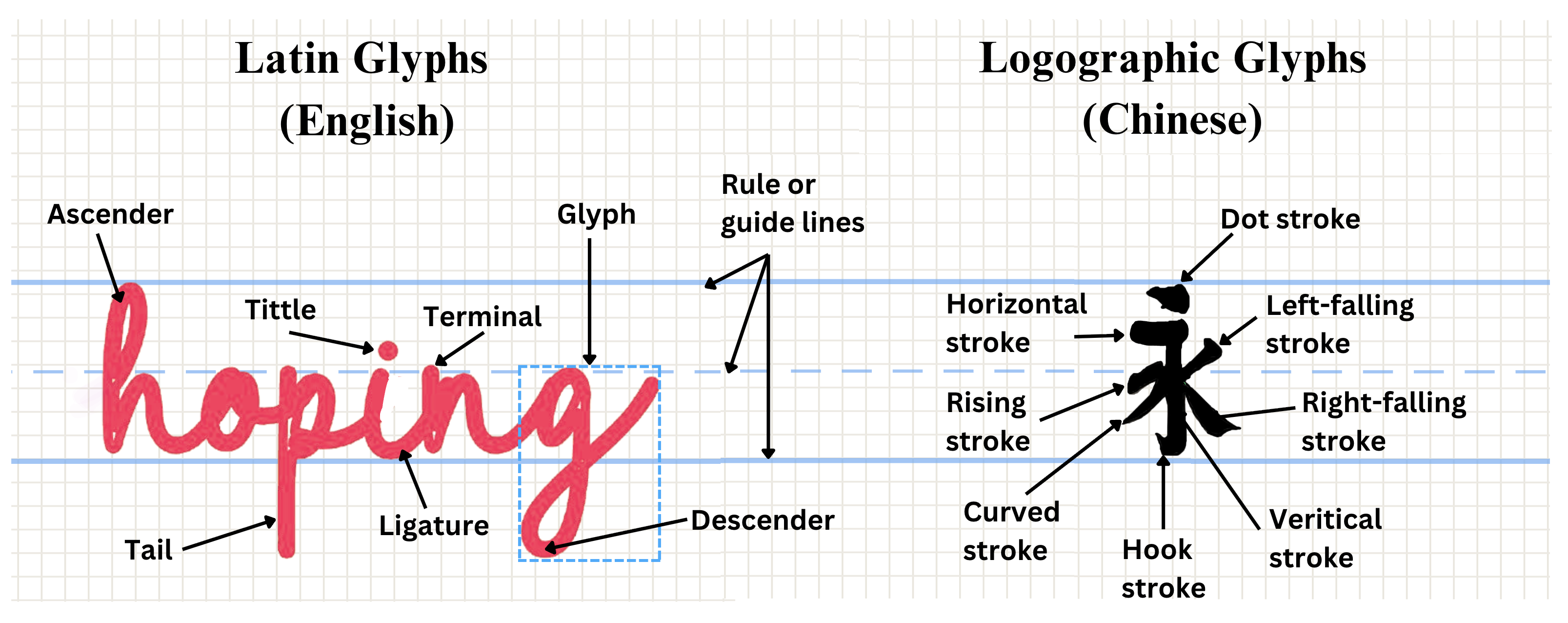}%
  \caption{Glossary of key parts of handwritten text, for both Latin and logographic scripts.}
  \label{fig:glossary}
\end{figure}

Using natural language to guide and discretely synthesize handwriting remains largely unexplored. Existing deep learning methods predominantly rely on reference images for style conditioning and generation, with limited control over style through natural language. As illustrated in \figref{fig:glossary}, handwritten text contains structural components that vary exponentially across scripts, requiring a synthesis method that can effectively leverage natural language guidance. Recent progress in LLM reasoning and agentic capabilities has made it feasible to utilize language in discrete handwriting generation. One of the earliest attempts at this is SketchAgent~\citep{Vinker2025}, which showed that an LLM-based agent can discretely synthesize hand-drawn sketches through iterative stroke-level actions guided by language. This indicates that large models may now possess the spatial and sequential reasoning needed to discretely synthesize diverse handwriting styles. Additionally, their broad multilingual and cross-domain capabilities enable generalization across diverse scripts and handwriting tasks, including handwritten maths and science expressions, with direct applications in education.

\begin{figure*}[!t]
  \centering
  \includegraphics[width=0.98\textwidth]{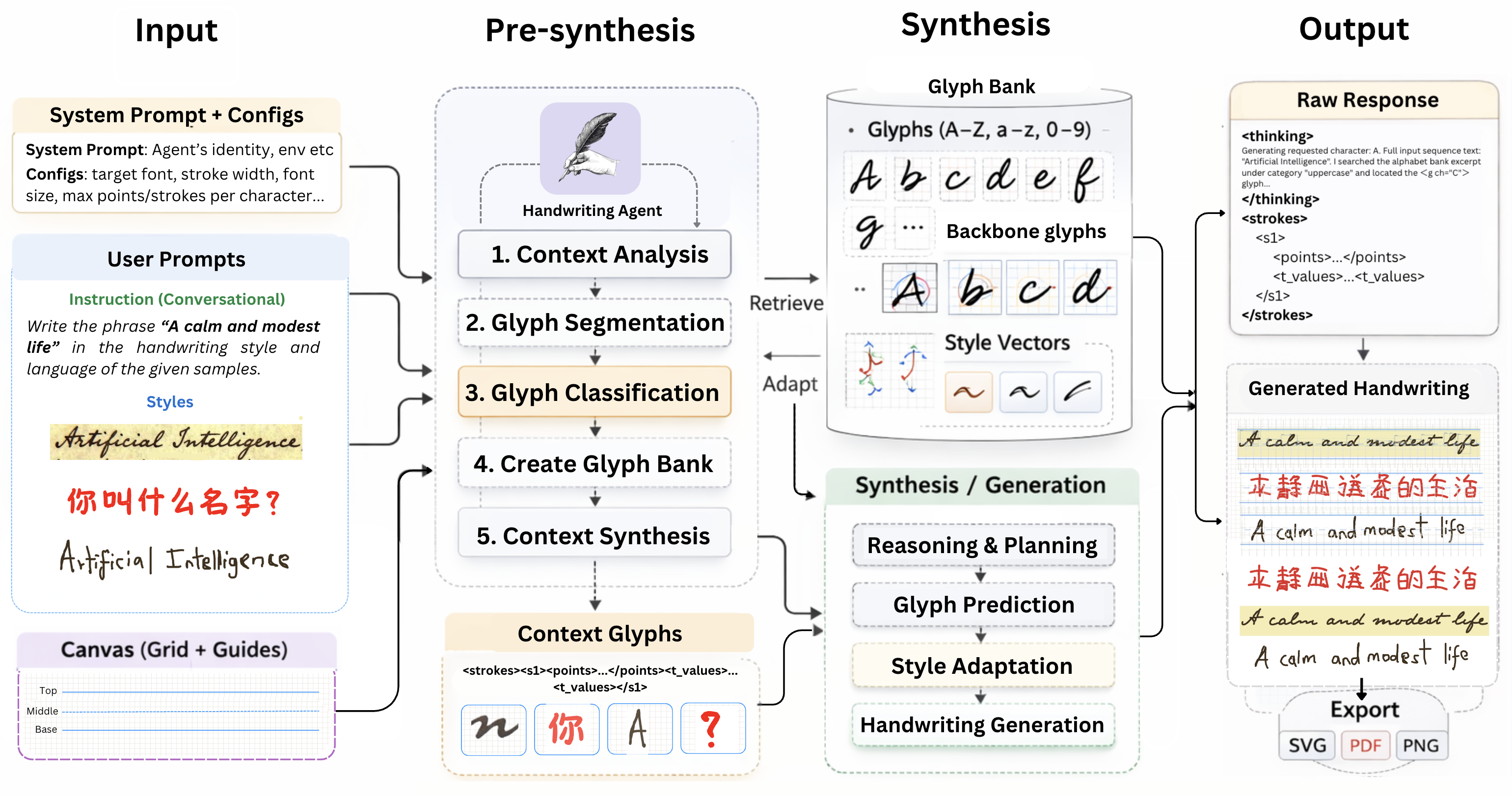}
  \caption{HandwritingAgent's framework. The handwriting task starts with a pre-synthesis stage, where the agent processes three main inputs: (1) a system prompt, (2) the user prompt (\ie handwritten texts in font, image or user handwriting strokes, writing configurations), and (3) a discrete grid canvas. This processing involves parsing input sequences, context analysis, glyph segmentation and classification, and creating a reusable reference glyph bank. The synthesis stage follows, wherein the agent performs multi-step reasoning and planning over its inputs and targets in natural language, then generates stroke sequences that closely reflect the geometric structure of the target handwriting style. Finally, generated strokes are processed, analysed and rendered onto the canvas as a sequence of pen strokes in SVG space.}
  \label{fig:architecture}
\end{figure*}

\section{Method}

As detailed in the framework shown in \figref{fig:architecture}, our primary objective is to develop an efficient, multi-task and multilingual handwriting synthesis agent that can discretely generate natural, controllable handwriting styles by reasoning over the geometric structure of written forms through natural language. To achieve this, HandwritingAgent is designed to support two core modes of operation:

\begin{enumerate}
  \item \textit{Conversational}: In this mode, the user interacts with the agent through iterative dialogue, giving instructions on what to write based on the given handwriting style.
  \item \textit{Non-conversational}: Here, the agent is given only the exact text to write in the handwriting style of the provided sample.
\end{enumerate}

Both modes accept reference handwriting samples in image and stroke formats, including user handwriting written directly on a canvas. The agent goes through three major stages when completing a given handwriting task: pre-synthesis, synthesis, and post-synthesis.

\subsection{Pre-synthesis}

Pre-synthesis begins by consolidating all inputs
into coherent reusable configurations that defines
both the stylistic behaviour of the target handwriting sequence and the geometric environment in
which the agent operates.

\subsubsection{Configurations}
\begin{figure}[H]
  \centering
  \includegraphics[width=0.95\columnwidth]{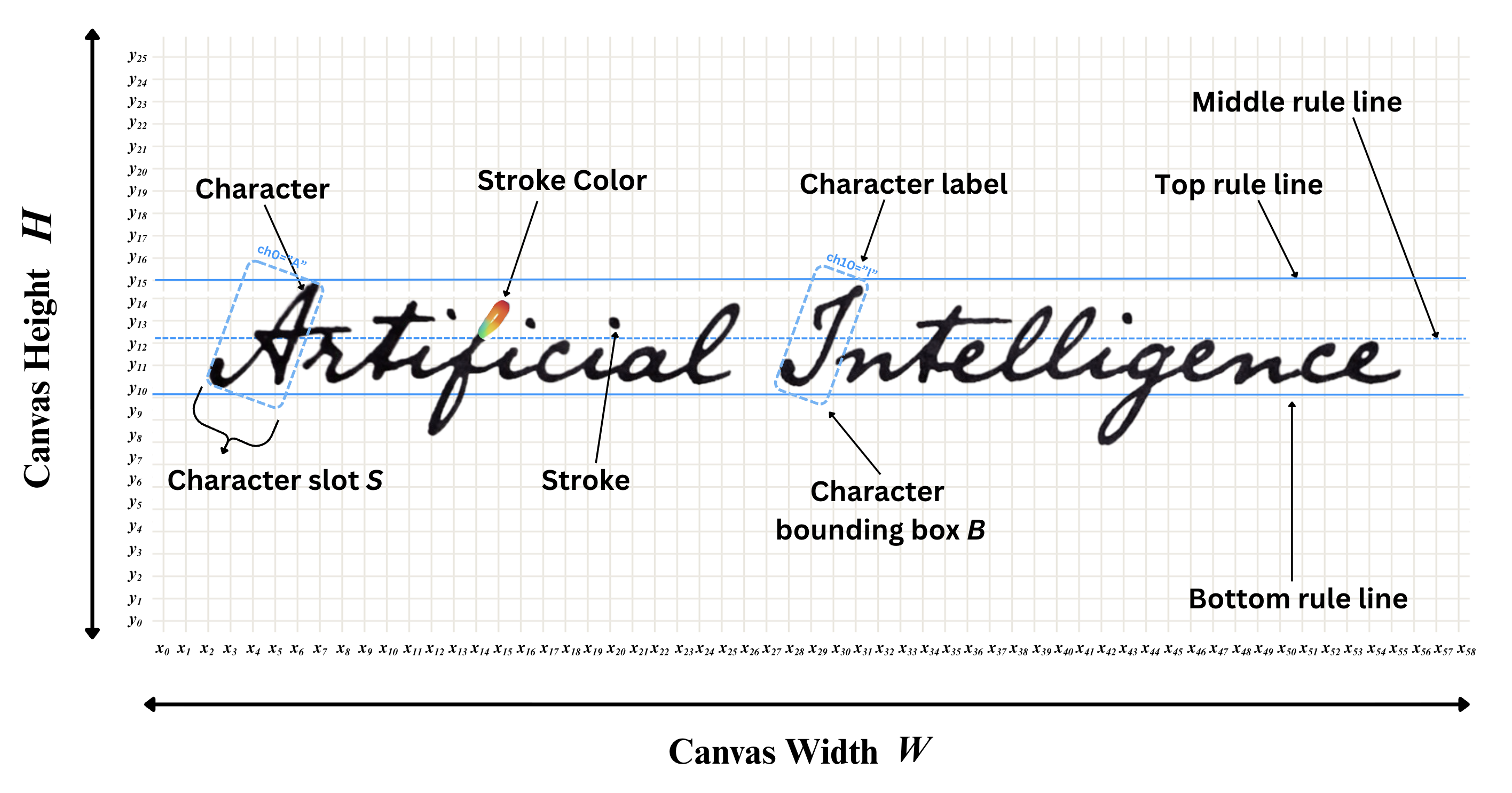}
  \caption{Illustration of the system canvas.}
  \label{fig:canvas}
\end{figure}

Typographical parameters cover the expressive range available to the agent: \emph{stroke width} sets the rendered ink thickness; limits on \emph{maximum strokes per character} and \emph{maximum points per stroke} regulate structural complexity and token usage; and optional Bézier control-point settings influence how the later curve-fitting stage interprets the geometry. Values for these configurations are dynamically adapted based on reasoning cues from the reference handwriting style and instruction. Spatial parameters define the canvas (illustrated in \figref{fig:canvas}) itself, expressed as a grid with a configurable number of rows and columns, along with letter spacing, font size, and color attributes that determine the visual rhythm and proportional scale of the final output. Together, these controls do not merely shape aesthetics, they determine the symbolic vocabulary, spatial resolution, and constraints under which the LLM must plan handwriting trajectories.

\subsubsection{Sample Handwriting Analysis}
When a handwriting request is made, all input data is first normalized into a unified representation that the LLM can iteratively reason on. User prompt, style configurations, and sample glyphs provide context for synthesis, while the grid canvas defines a shared coordinate space in which letterforms, stroke trajectories, and spatial guiding lines can be represented and reasoned over consistently. The style image is converted to grayscale, filtered to isolate ink regions or stroke paths, segmented and classified into \textit{character} units. The resulting processed context data is serialized as structured XML containing global metadata for reasoning. Because these representations are explicit rather than latent, each component can be inspected and interpreted, forming a stable substrate for precise reasoning during synthesis.

\subsubsection{Glyph Annotation}
In order to improve synthesis, the sample handwriting is first automatically segmented and labelled before any generation is attempted. For this, we perform character or word-level segmentation and classification by leveraging the underlying LLM as an OCR model, applying image-processing heuristics for character extraction with LLM-based annotation. However, this does not often yield accurate results, especially for cursive handwriting, where glyphs inherently have stroke continuity across characters. To enhance performance, we first have the agent apply adjustments to the coordinates and orientations of segment boundaries prior to labeling. We also add an optional human-in-the-loop annotation stage, in which the user edits the annotated glyphs directly on a canvas, if needed. Corrected annotations are then persisted as structured XML context data used in creating glyph banks, predicting missing glyphs, and in the downstream handwriting synthesis.

\subsubsection{Glyph Bank}
\begin{figure}[t]
  \centering
  \includegraphics[width=\columnwidth]{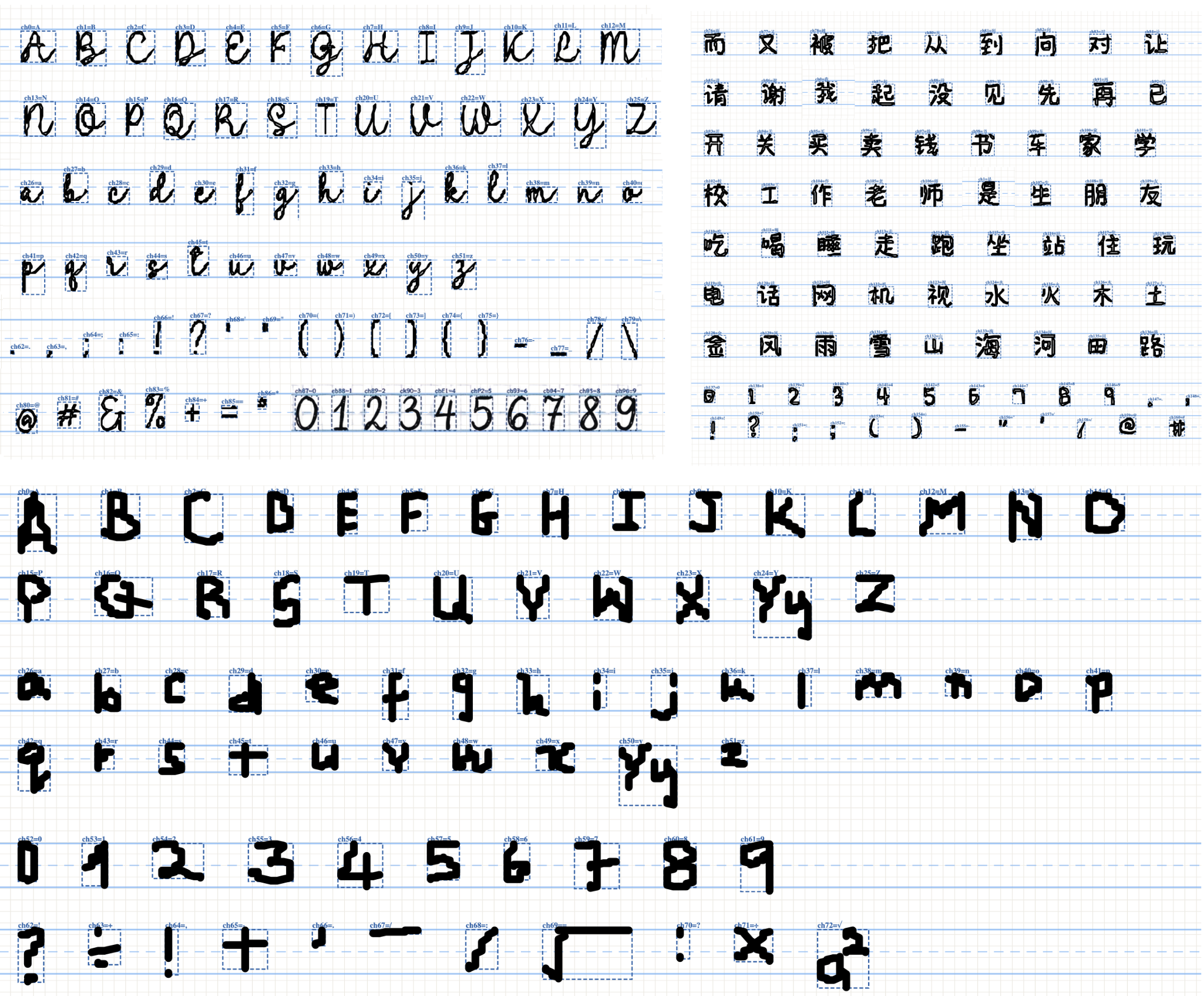}%
  \caption{Sample glyph banks. Top-left: glyph bank for when input is conditioned on a Latin style (English). Top-right: glyph bank for when input is conditioned on a logographic style (Mandarin). Bottom: glyph bank for when input is conditioned on a user's handwriting written directly on the canvas.}
  \label{fig:glyph-bank}
\end{figure}

Following annotation, the agent constructs a glyph bank, which serves as a structural and stylistic reference for handwriting generation in the instructed or prompted language. The glyph bank (see examples in \figref{fig:glyph-bank}) is derived from processed handwriting samples, where extracted strokes are normalized, aligned, and distilled into canonical representations that preserve stylistic characteristics. During synthesis, the agent uses this bank to compare character forms, infer missing structures, and maintain consistency across newly generated outputs. In this way, the glyph bank provides an explicit stylistic memory that grounds the LLM in preserving fidelity to the reference handwriting across different words, symbols, and scripts.

\subsection{Synthesis}

\subsubsection{Reasoning and Planning}
\begin{figure}[b]
  \centering
  \includegraphics[width=\columnwidth]{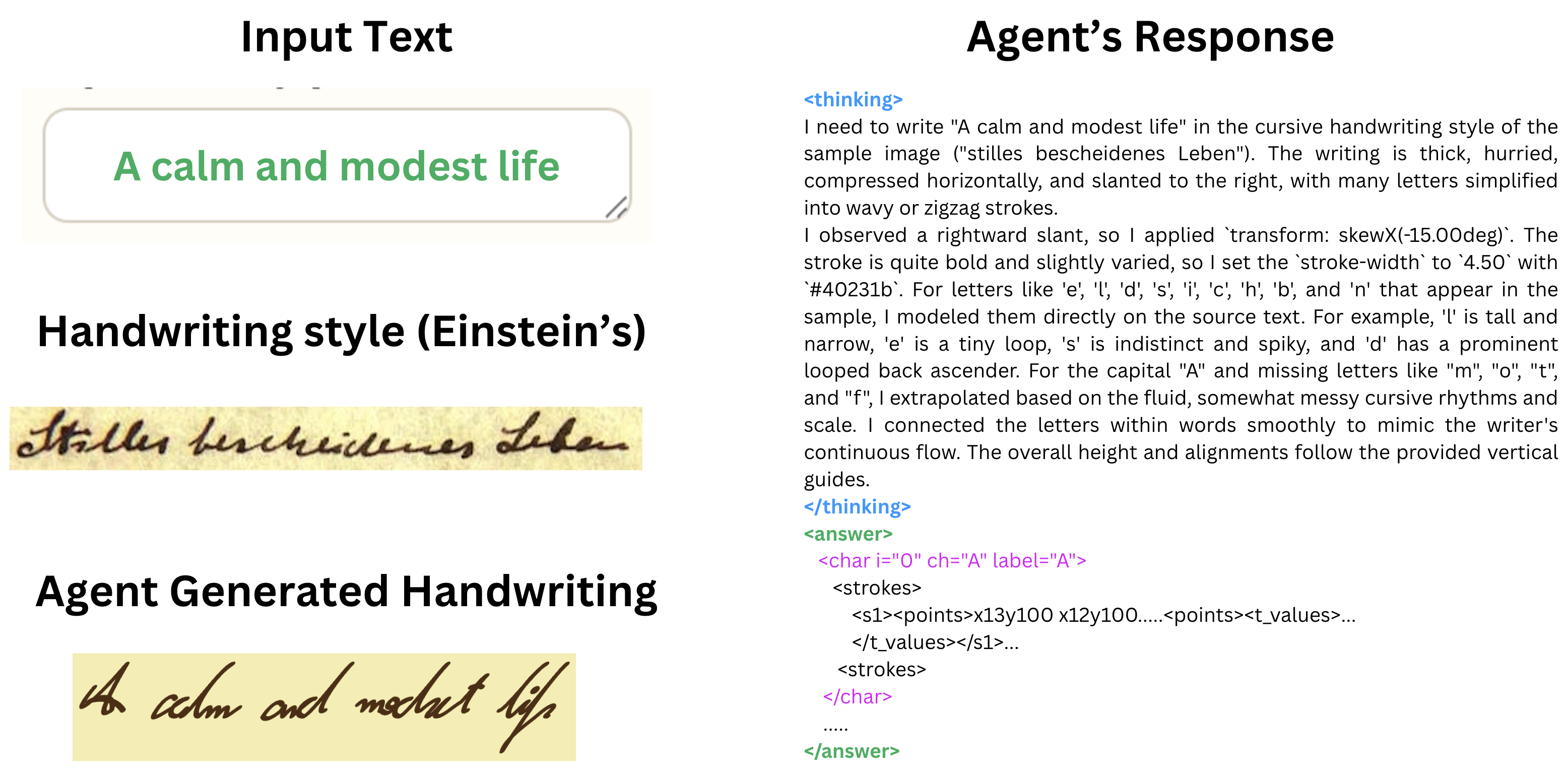}%
  \caption{Sample generated handwriting, showing the agent's explicit chain-of-thought process.}
  \label{fig:thinking-trace}
\end{figure}

During synthesis, the agent functions as a geometric reasoner and planner that starts by analysing the observed writing style in the given sample. This includes understanding the written content, identifying stroke patterns, alignment, curvature tendencies, spacing, and other structural dependencies. Based on this analysis, it then formulates a step-by-step plan for how to adapt each target character to the observed handwriting style. This plan accounts for the full target text sequence, reusable features in the sample, and specifies how these cues can be utilized during synthesis. As shown in \figref{fig:thinking-trace}, the agent exposes this intermediate chain-of-thought in \texttt{<thinking>...</thinking>} tags, which is followed by generated SVG stroke sequences in \texttt{<answer><char><strokes>...} style tags. This explicit reasoning and planning strategy is especially important when predicting missing glyphs that do not directly appear in the reference sample, since the agent must infer plausible forms from contextual and stylistic cues rather than just copying observed evidence.

\subsubsection{Glyph Prediction and Style Adaption}
The agent performs glyph prediction and style adaptation through reasoning-informed predictive style transfer. The reference handwriting is first segmented into character- or word-level crops while preserving geometric metadata such as position, stroke width, and stroke trajectory. Each target character is then matched against this handwritten context bank to identify the most suitable exemplar. When an exact glyph is available, it provides the primary basis for synthesis. For characters absent from the reference sample, the agent retrieves the corresponding form from a language-specific backbone glyph bank and adapts it to the observed handwriting style. This adaptation is carried out through iterative reasoning steps that regulates key writing-form attributes, including slant, stroke proportions, spacing, curvature, and ligature patterns.

\subsubsection{Generation}
Generation is carried out at the character level, with the agent synthesizing each character as an ordered sequence of stroke-point coordinates on the canvas. Each stroke-point sequence has associated temporal values that permit dynamic writing and smooth fitting on a cubic Bézier curve. A cubic Bézier curve is a polynomial with four control points $P_{0}, P_{1}, P_{2}, P_{3}$, expressed as
\begin{equation}
  \begin{aligned}
    B(t) ={}& (1 - t)^{3}P_{0} + 3(1 - t)^{2}tP_{1} \\
            & + 3(1 - t)t^{2}P_{2} + t^{3}P_{3}, \quad 0 \leq t \leq 1.
  \end{aligned}
\end{equation}
Here $P_{0}$ and $P_{3}$ denote the start and end points, and $P_{1}$ and $P_{2}$ are control points that determine the curvature of the stroke. Thus, for a sequence of strokes $S = \{S_{0}, S_{1}, \ldots, S_{i}\}$, each stroke $S_{i}$ is represented as an ordered set of $n$ cell coordinates sampled along its corresponding Bézier curve:
\begin{equation}
  S_{i} = \left\{\left(p_{k}^{(i)}, t_{k}^{(i)}\right)\right\}_{k=1}^{n}, \quad 0 \leq t_{k}^{(i)} \leq 1.
\end{equation}
where $p_{k}^{(i)} = \left(x_{k}^{(i)}, y_{k}^{(i)}\right)$ denotes the $k$-th sampled cell coordinate and $t_{k}^{(i)}$ specifies its relative position along the curve. In practice, the agent emits strokes as structured XML of the form \texttt{<strokes><s1><points>...}, which are then parsed into ordered point sequences with their associated timestamps, mapped to grid coordinates, assembled into an InkDocument, smoothed using Bézier curves, and finally converted into SVG \texttt{<path>} elements for rendering.

\subsection{Post-synthesis}

Once synthesis is complete, generated stroke sequences are processed and converted into SVG format, with geometric and temporal data preserved for dynamic writing on the canvas.

\subsubsection{Writing Dynamics}
\begin{figure}[h]
  \centering
  \includegraphics[width=\columnwidth]{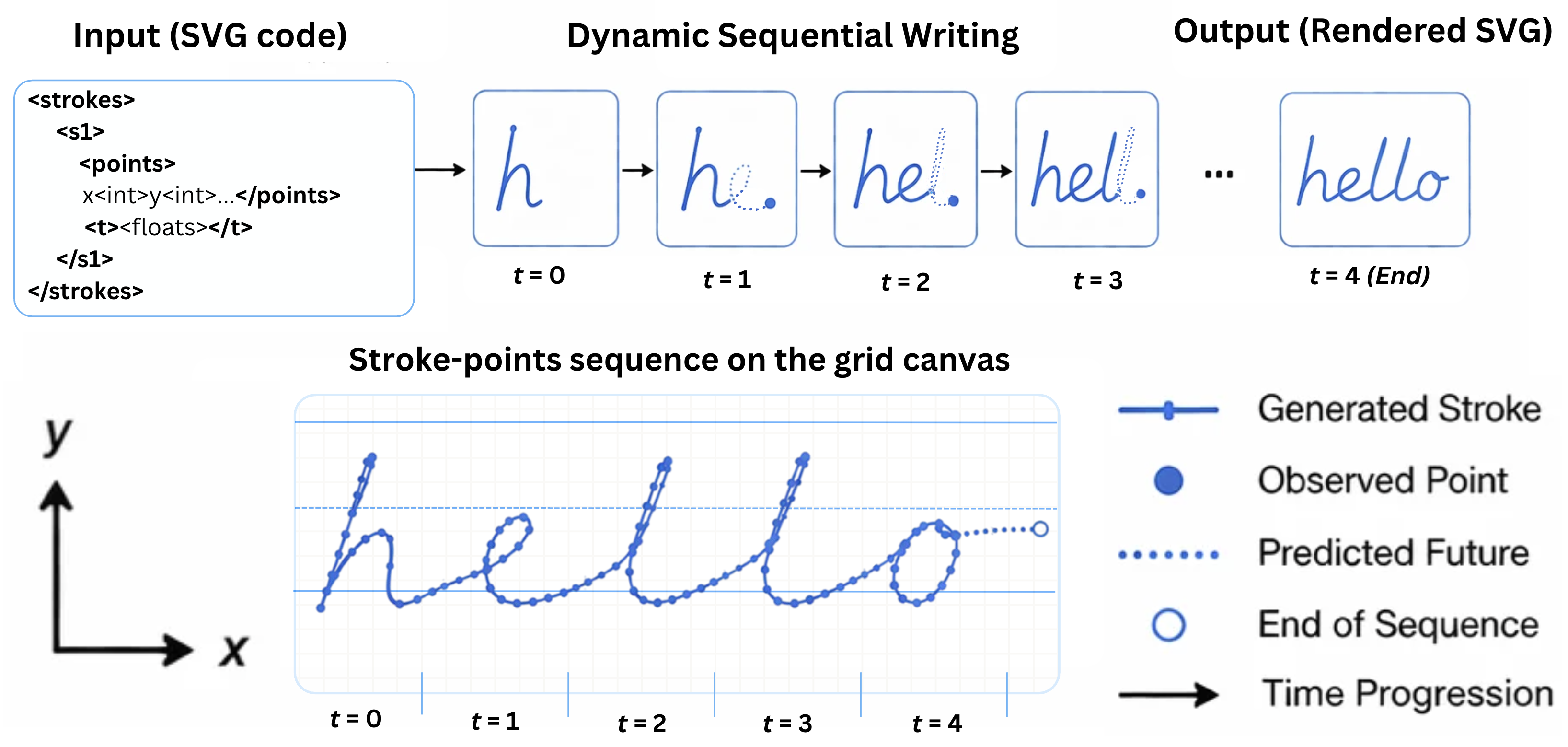}%
  \caption{Illustration of HandwritingAgent's writing dynamics.}
  \label{fig:dynamics}
\end{figure}

As illustrated in \figref{fig:dynamics}, we model the agent's writing dynamics as a natural, sequential handwriting process in which each generated glyph is transformed into an ordered stroke sequence, resampled into its bounded slot, and then dynamically rendered stroke by stroke according to its geometric and temporal configurations. These configurations include key writing dependencies such as stroke width, point radius, and timestamps, which together allow the agent to preserve continuity, stroke progression, and stylistic motion during synthesis, so that dynamically rendered glyph sequences better reflect the properties of real human handwriting.

\section{Experiments}

We conducted three experiments to address the following research questions for real-world use-case scenarios:

\begin{enumerate}
  \item RQ1: Handwriting Imitation. How well can the agent imitate a target writer's handwriting style when given only a small reference sample of it?
  \item RQ2: Multilingual Handwriting Synthesis. How effectively can the agent accurately recognize, translate and synthesize handwritten text in a specified language? 
  \item RQ3: Generating Handwritten Maths and Science Expressions. To what extent can the agent imitate complex handwritten maths and science expressions or diagrams with sufficient fidelity to be useful in educational applications?
\end{enumerate}

\subsection{Experiment Setup}
\label{sec:experiment-setup}

\subsubsection{Implementation Details}
We used GPT 5.6 Sol for all experiments with \textit{thinking} enabled. For handwriting imitation (RQ1), we evaluate the system on 250 samples from the IAM handwriting dataset (both lines and words, 50 writers each) and benchmark against the following state-of-the-art methods: HWT, DiffPen, Emuru, Eruku, One-DM, and VATr++. For multilingual handwriting synthesis (RQ2), we experiment on Chinese and English scripts, using 250 samples per language from CASIA-HWDB2.1~\citep{Liu2011} and the IAM dataset~\citep{Marti2002}, respectively. Similarly, for handwritten maths expressions (RQ3), we evaluate our approach on 250 samples from CROHME 2014~\citep{Mouchere2014}, assessing it against the HMEG~\citep{Chen2024} and FormulaGAN~\citep{Springstein2021} baselines. Finally, for scientific expressions (RQ3), we consider two science subjects: chemistry, evaluated on 250 samples of EDU-CHEMC~\citep{Hu2023}, and physics, evaluated on 250 curated samples from handwritten physics lecture notes~\citep{Richmond2001}. To avoid prohibitive costs during experiments and fit the available API token budget cap, we evaluated each benchmark using a fixed subset of 250 samples\footnote{The sampling code per benchmark is available in the project's open-source repository for reproducibility.} across all experiments. To ensure fair metric comparisons against baseline methods, we converted HandwritingAgent's native SVG outputs into raster images (\texttt{.png}), with the raw data saved for further validation.

\begin{figure*}[!t]
  \centering
  \includegraphics[width=1.0\textwidth]{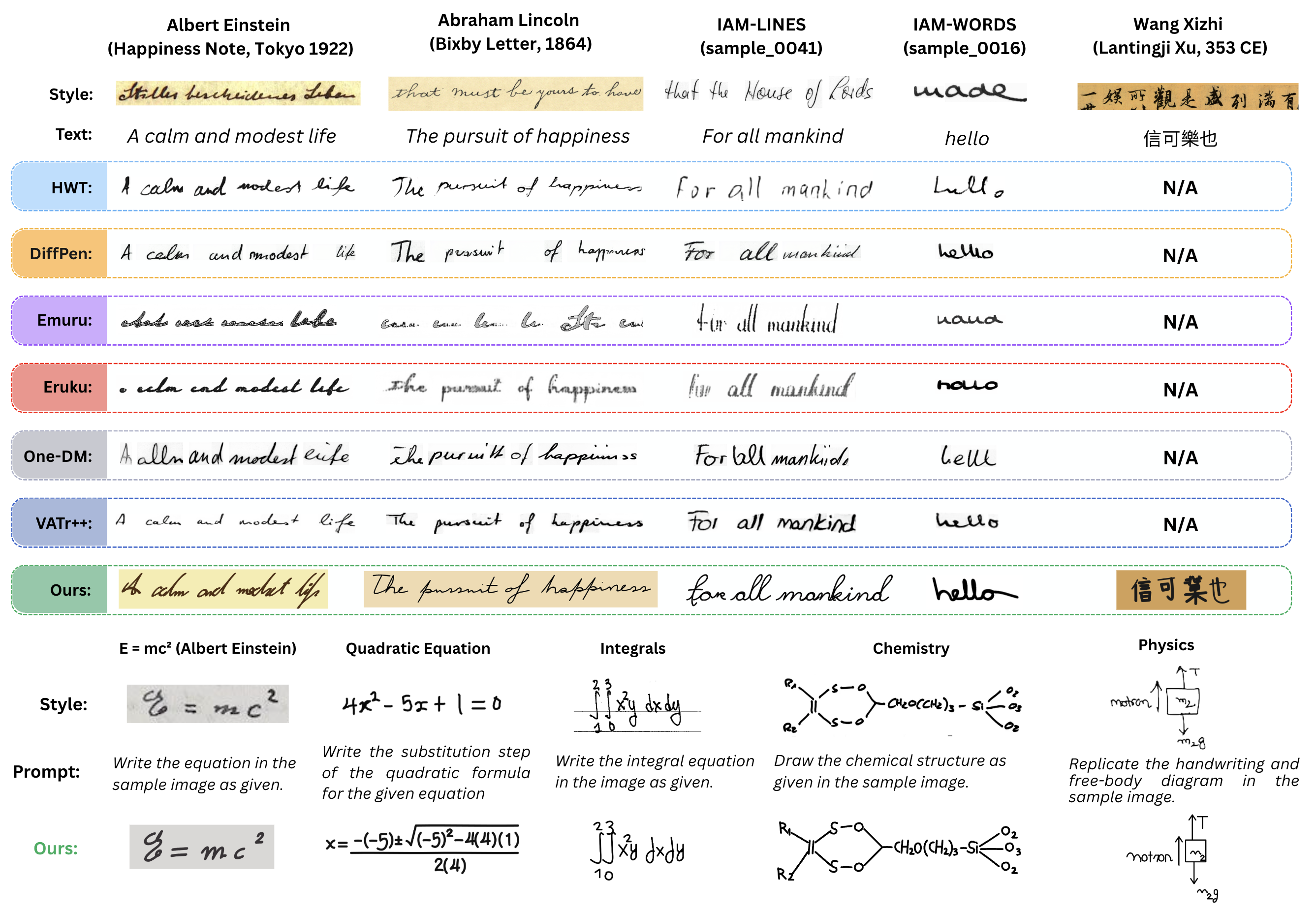}%
  \caption{(Top) Qualitative results across diverse handwriting styles, comparing HandwritingAgent against state-of-the-art handwriting synthesis models. Note that HandwritingAgent dynamically adapts background color to match the given reference style. (Bottom) Qualitative results on synthesizing handwritten maths and science expressions. As shown in the quadratic equation sample, the agent can also follow explicit instructions on what to write or draw based on a given style reference.}
  \label{fig:qualitative}
\end{figure*}

\begin{table}[t]
  \centering
  \small
  \setlength{\tabcolsep}{1pt}
  \begin{tabular}{@{}lcccc@{}}
    \toprule
    & \multicolumn{4}{c}{\textbf{IAM Words ($N=250$)}} \\
    \cmidrule(lr){2-5}
    \textbf{Method} & \textbf{SSIM}$\uparrow$ & \textbf{CER}$\downarrow$ & \textbf{FID}$\downarrow$ & \textbf{HWD}$\downarrow$ \\
    \midrule
    HWT & 0.37 & \underline{0.30} & 104.98 & 2.30 \\
    DiffPen & \underline{0.58} & 0.37 & 90.17 & 1.93 \\
    Emuru & 0.43 & 0.69 & 117.77 & \underline{1.62} \\
    Eruku & 0.54 & 1.09 & 119.47 & 2.45 \\
    One-DM & 0.38 & 0.41 & 79.96 & 2.44 \\
    VATr++ & 0.38 & \textbf{0.24} & \underline{76.28} & 2.65 \\
    \textbf{Ours} & \textbf{0.63} & 0.43 & \textbf{70.07} & \textbf{1.36} \\
    \bottomrule
  \end{tabular}
  \caption{(RQ1) Quantitative results on 250 samples of the IAM-Words dataset.}
  \label{tab:rq1-word}
\end{table}

\begin{table}[t]
  \centering
  \small
  \setlength{\tabcolsep}{1pt}
  \begin{tabular}{@{}lcccc@{}}
    \toprule
    & \multicolumn{4}{c}{\textbf{IAM Lines ($N=250$)}} \\
    \cmidrule(lr){2-5}
    \textbf{Method} & \textbf{SSIM}$\uparrow$ & \textbf{CER}$\downarrow$ & \textbf{FID}$\downarrow$ & \textbf{HWD}$\downarrow$ \\
    \midrule
    HWT & 0.52 & 0.63 & 91.46 & 2.17 \\
    DiffPen & 0.43 & 0.56 & 90.17 & \underline{1.93} \\
    Emuru & \underline{0.72} & 0.73 & \underline{65.90} & 1.97 \\
    Eruku & 0.57 & 0.45 & 75.38 & 2.09 \\
    One-DM & 0.48 & \underline{0.39} & 103.85 & 2.68 \\
    VATr++ & 0.58 & 0.72 & 93.74 & 2.18 \\
    \textbf{Ours} & \textbf{0.75}  & \textbf{0.36} & \textbf{55.25} & \textbf{1.32} \\
    \bottomrule
  \end{tabular}
  \caption{(RQ1) Quantitative results on 250 samples of the IAM Lines dataset.}
  \label{tab:rq1-line}
\end{table}

\subsubsection{Evaluation Metrics}
For handwriting imitation and multilingual handwriting synthesis, we report Structural Similarity Measure (SSIM)~\citep{Wang2004}, Character Error Rate difference (CER)~\citep{Pippi2025}, Fréchet Inception Distance (FID)~\citep{Heusel2017}, and Handwriting Distance (HWD)~\citep{Pippi2023}. For handwritten maths and science expressions, we report Expression Recognition Rate (ExpRate) and Word Error Rate (WER), in addition to SSIM and CER. Lower values of FID, CER, HWD, and WER indicate better realism of the synthesized images, while greater values of SSIM and ExpRate means higher similarity between synthesized and real images.

\subsection{Results}

\subsubsection{Handwriting Imitation (RQ1)}
RQ1 results show that HandwritingAgent per-

\noindent forms strongest on visual and structural imitation, while readability remains competitive with the best baselines. As reported in \tabref{tab:rq1-word}, on IAM Words, the agent records leading SSIM, FID, and HWD scores of 0.63, 70.07, and 1.36, respectively, indicating stronger preservation of handwriting form, visual distribution, and writer-style similarity. VATr++ attains the leading CER score, while HandwritingAgent records 0.43, reflecting a remaining gap in character-level readability.

On IAM Lines (reported in \tabref{tab:rq1-line}), HandwritingAgent achieves the best overall performance, with the highest SSIM and the lowest CER, FID, and HWD scores, indicating stronger structural fidelity, readability, distributional similarity, and writer-style consistency for longer handwriting sequences. Qualitative comparisons in \figref{fig:qualitative} further validate these results, showing more consistent preservation of sample-specific stroke shape, spacing, background cues, and match to task requirements across word-level, line-level, multilingual, and STEM handwriting cases.

\subsubsection{Multilingual Handwriting Synthesis (RQ2)}
\begin{figure}[t]
  \centering
  \includegraphics[width=\columnwidth]{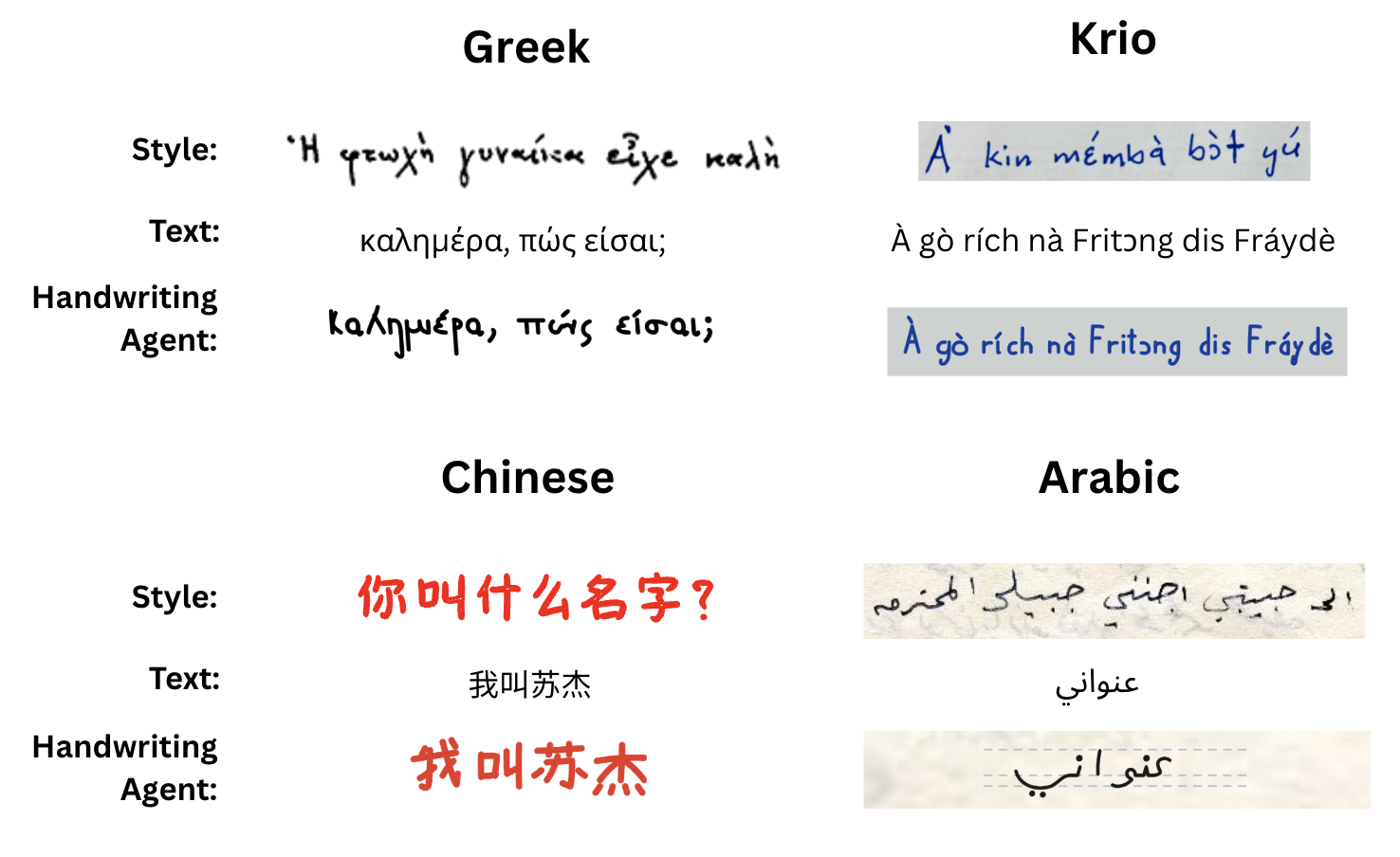}%
  \caption{Qualitative multilingual results across Greek, Krio, Chinese, and Arabic scripts.}
  \label{fig:additional-multilingual}
\end{figure}

Synthesis results across additional languages further demonstrate the agent's ability to generalize beyond Latin scripts, provided that the underlying LLM supports the target languages. As reported in \tabref{tab:rq2-chinese}, HandwritingAgent achieves the lowest FID on CASIA-HWDB2.1, while DiffBrush~\citep{Dai2025DiffBrush} obtains the better HWD, reflecting complementary strengths in visual fidelity and style similarity. As shown in \figref{fig:additional-multilingual}, this capability extends qualitatively to Greek, Krio, and Arabic, where the agent preserves the supplied handwriting style while generating text across distinct scripts and writing directions.

\begin{table}[H]
  \centering
  \small
  \setlength{\tabcolsep}{3pt}
  \renewcommand{\arraystretch}{1.12}
  \begin{tabular}{@{}
    >{\raggedright\arraybackslash}m{0.22\columnwidth}
    >{\centering\arraybackslash}m{0.13\columnwidth}
    >{\centering\arraybackslash}m{0.13\columnwidth}
    >{\centering\arraybackslash}m{0.46\columnwidth}
  @{}}
    \toprule
    \multicolumn{4}{c}{\textbf{Chinese: CASIA-HWDB2.1 ($N=250$)}} \\
    \cmidrule(lr){1-4}
    \textbf{Method}
      & \textbf{FID}$\downarrow$
      & \textbf{HWD}$\downarrow$
      & ~\includegraphics[
          width=\linewidth,
          height=0.11\textheight,
          keepaspectratio
        ]{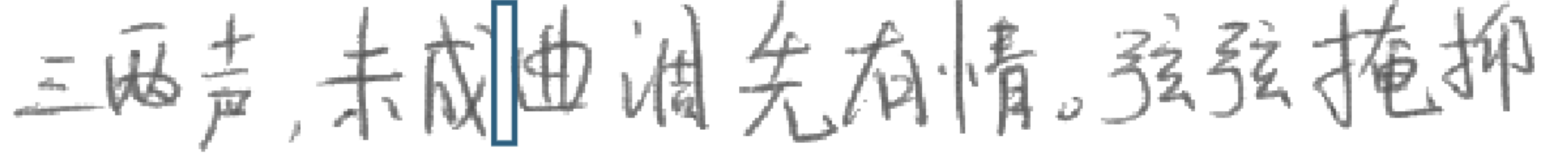} \\
    \midrule
    DiffBrush\footnotemark
      & \underline{7.84}
      & \textbf{0.73}
      & \includegraphics[
          width=\linewidth,
          height=0.11\textheight,
          keepaspectratio
        ]{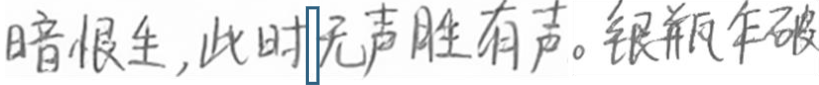} \\
    \textbf{Ours}
      & \textbf{0.02}
      & \underline{1.12}
      & \includegraphics[
          width=\linewidth,
          height=0.20\textheight,
          keepaspectratio
        ]{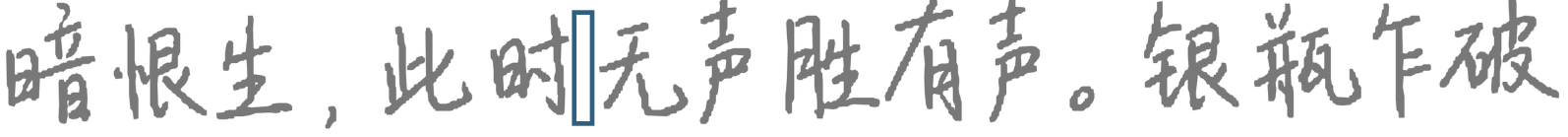} \\
    \bottomrule
  \end{tabular}
  \caption{(RQ2) Quantitative and qualitative comparison with DiffBrush on 250 CASIA-HWDB2.1 Chinese text-line samples.}
  \label{tab:rq2-chinese}
\end{table}

\footnotetext{DiffBrush reports quantitative and qualitative results for Chinese handwriting synthesis but does not open-source the model to reproduce them. We attempted training with their released codebase, but performance was not reliable enough for comparison; therefore, we compare against their reported baselines.}

\subsubsection{Generating Handwritten Maths and Science Expressions (RQ3)}

Synthesizing structurally complex maths and 

\noindent science expressions introduces additional difficulty, as generated samples must be legible, spatially valid, and symbolically interpretable. As shown in \tabref{tab:rq3-math}, HandwritingAgent achieves the strongest results across all four metrics on CROHME 2014, with the highest SSIM and ExpRate and lowest CER and WER relative to the HMEG and FormulaGAN baselines. This indicates that the generated mathematical expressions preserve both visual structure and symbolic recognizability, although exact reconstruction remains challenging. The science-expression results in \tabref{tab:rq3-science} further show a distinction between symbolic and diagrammatic content. Chemistry expressions remain relatively stable, while physics samples are more fragile, suggesting that diagrams, labels, arrows, and spatial annotations impose greater coordination demands on the agent.

\begin{table}[H]
  \centering
  \small
  \setlength{\tabcolsep}{1pt}
  \begin{tabular}{@{}lcccc@{}}
    \toprule
    \multicolumn{5}{c}{\textbf{Math: CROHME 2014 ($N=250$)}} \\
    \cmidrule(lr){1-5}
    \textbf{Method} & \textbf{SSIM}$\uparrow$ & \textbf{CER}$\downarrow$ & \textbf{WER}$\downarrow$ & \textbf{ExpRate}$\uparrow$ \\
    \midrule
    HMEG & \underline{0.62} & \underline{0.80} & 0.75 & 0.00 \\
    FormulaGAN & 0.39 & 0.85 & \underline{0.83} & \underline{0.02} \\
    \textbf{Ours} & \textbf{0.88} & \textbf{0.53} & \textbf{0.52} & \textbf{0.12} \\
    \bottomrule
  \end{tabular}
  \caption{(RQ3) Comparative results with HMEG and FormulaGAN for mathematics expressions on 250 samples from the CROHME 2014 dataset.}
  \label{tab:rq3-math}
\end{table}

\begin{table}[!htbp]
  \centering
  \small
  \setlength{\tabcolsep}{1pt}
  \begin{tabular}{@{}lcccc@{}}
    \toprule
    & \multicolumn{4}{c}{\textbf{Science ($N=250$)}} \\
    \cmidrule(lr){2-5}
    \textbf{Domain} & \textbf{SSIM}$\uparrow$ & \textbf{CER}$\downarrow$ & \textbf{WER}$\downarrow$ & \textbf{ExpRate}$\uparrow$ \\
    \midrule
    \shortstack[l]{Chemistry} & \textbf{0.68} & \textbf{0.01} & 0.91 & 0.27 \\
    \shortstack[l]{Physics} & 0.54 & 0.69 & \textbf{0.76} & \textbf{0.80} \\
    \bottomrule
  \end{tabular}
  \caption{(RQ3) Quantitative results of HandwritingAgent synthesizing handwritten science expressions on 250 samples each from the EDU-CHEMC and Physics 311 lecture notes datasets.}
  \label{tab:rq3-science}
\end{table}

\subsection{Ablations}

To assess the effect of reasoning on synthesis, we conduct ablation studies on the IAM dataset by comparing agent outputs generated with and without the underlying LLM’s thinking mode. This setting allows us to isolate whether explicit reasoning improves the agent’s ability to interpret reference handwriting, plan stroke-level actions, and preserve writer-specific style during synthesis. Both settings are evaluated using the same metrics described in \secref{sec:experiment-setup} to ensure comparability across IAM Words and IAM Lines tasks.

As shown in \tabref{tab:ablation-iam-word} and \tabref{tab:ablation-iam-line}, enabling \textit{thinking} improves performance across all metrics on both IAM Words and IAM Lines. SSIM on IAM Lines is a tie, indicating that the agent can preserve strong structural similarity even without explicit reasoning; however, \textit{thinking} provides additional gains in style adaptation, accuracy, and fidelity. Qualitative results in \tabref{tab:ablation-qualitative} further validate this claim with consistent stroke fidelity, spacing, and style adaptation when \textit{thinking} is enabled.

\begin{table}[H]
  \centering
  \small
  \setlength{\tabcolsep}{2pt}
  \renewcommand{\arraystretch}{1.08}
  \begin{tabular}{@{}
    >{\raggedright\arraybackslash}m{0.28\columnwidth}
    *{4}{>{\centering\arraybackslash}m{0.16\columnwidth}}
  @{}}
    \toprule
    & \multicolumn{4}{c}{\textbf{Ablation: IAM Words ($N=250$)}} \\
    \cmidrule(lr){2-5}
    & \textbf{SSIM}$\uparrow$
    & \textbf{CER}$\downarrow$
    & \textbf{FID}$\downarrow$
    & \textbf{HWD}$\downarrow$ \\
    \midrule
    w/ thinking    & \textbf{0.63} & \textbf{0.43} & \textbf{70.07}  & \textbf{1.35} \\
    w/o thinking   & 0.38 & 1.31 & 112.64 & 3.69 \\
    \bottomrule
  \end{tabular}
  \caption{Quantitative ablation results on 250 samples of IAM Words with \textit{thinking} enabled and disabled.}
  \label{tab:ablation-iam-word}
\end{table}

\begin{table}[H]
  \centering
  \small
  \setlength{\tabcolsep}{2pt}
  \renewcommand{\arraystretch}{1.08}
  \begin{tabular}{@{}
    >{\raggedright\arraybackslash}m{0.28\columnwidth}
    *{4}{>{\centering\arraybackslash}m{0.16\columnwidth}}
  @{}}
    \toprule
    & \multicolumn{4}{c}{\textbf{Ablation: IAM Lines ($N=250$)}} \\
    \cmidrule(lr){2-5}
    & \textbf{SSIM}$\uparrow$
    & \textbf{CER}$\downarrow$
    & \textbf{FID}$\downarrow$
    & \textbf{HWD}$\downarrow$ \\
    \midrule
    w/ thinking  & \textbf{0.75} & \textbf{0.36} & \textbf{55.25}  & \textbf{1.32} \\
    w/o thinking & \textbf{0.75} & 0.46 & 102.39 & 1.97 \\
    \bottomrule
  \end{tabular}
  \caption{Quantitative ablation results on 250 samples of IAM Lines with \textit{thinking} enabled and disabled.}
  \label{tab:ablation-iam-line}
\end{table}

\begin{table}[t]
  \centering
  \small
  \setlength{\tabcolsep}{1pt}
  \renewcommand{\arraystretch}{1.14}
  \begin{tabular}{@{}>{\raggedright\arraybackslash}m{0.26\columnwidth} >{\centering\arraybackslash}m{0.35\columnwidth} >{\centering\arraybackslash}m{0.35\columnwidth}@{}}
    \toprule
    & \multicolumn{2}{c}{\textbf{Qualitative Ablation}} \\
    \cmidrule(lr){2-3}
    & \textbf{IAM Words} & \textbf{IAM Lines} \\
    \midrule
    & \textbf{Style:}~\ablationstyleimage{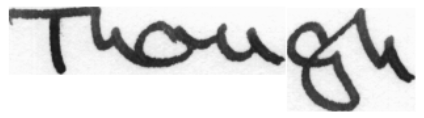}
      & \textbf{Style:}~\ablationstyleimage{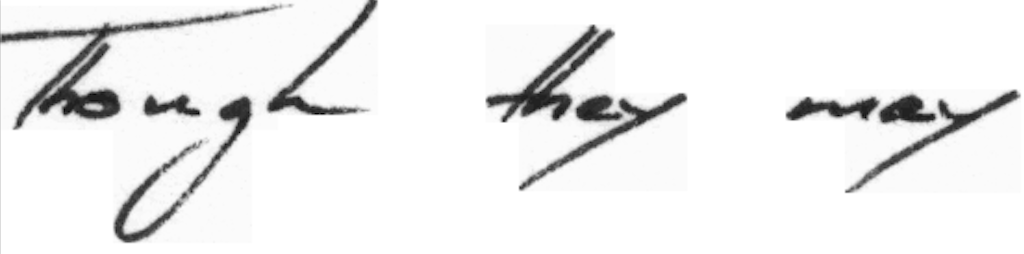} \\
    \midrule
    w/ thinking & \ablationresultimage{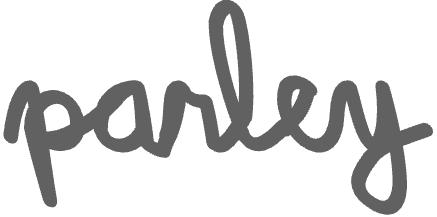} & \ablationresultimage{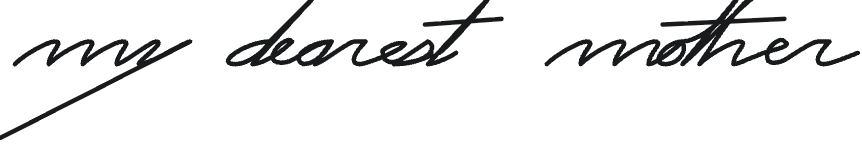} \\
    w/o thinking & \ablationresultimage{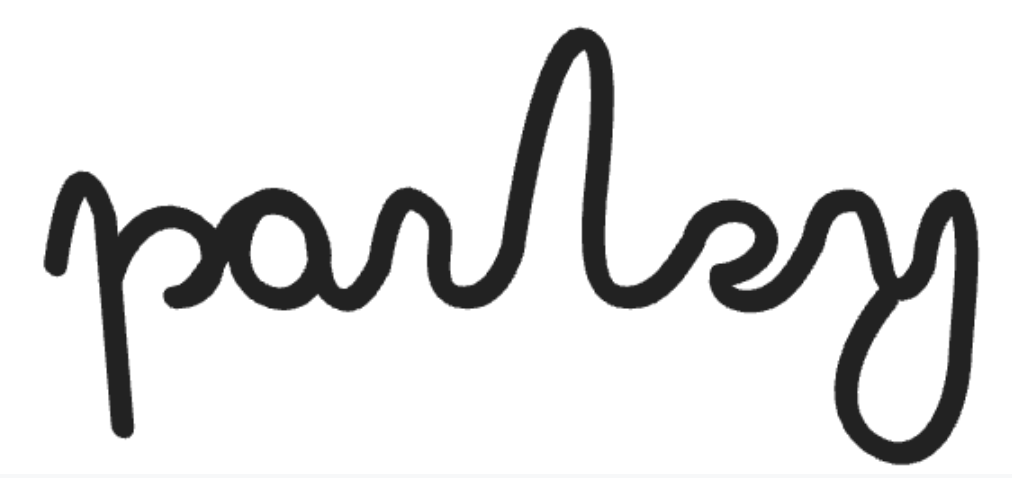} & \ablationresultimage{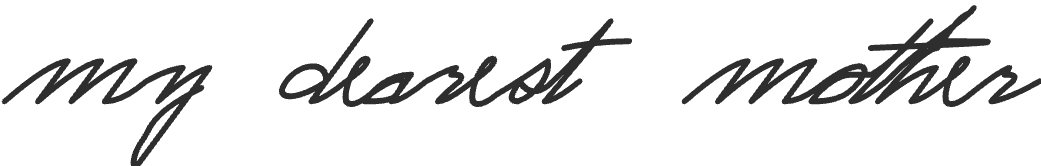}\\
    \bottomrule
  \end{tabular}
  \caption{Qualitative ablation results with \textit{thinking} mode enabled and disabled on representative IAM Words and IAM Lines samples.}
  \label{tab:ablation-qualitative}
\end{table}

\section{Limitations and Ethical Considerations}
We identify several limitations and ethical concerns that may inform future work. First, performance depends on the reasoning and multimodal capabilities of the underlying LLM, including the scripts and linguistic patterns it can reliably process. Second, quality varies across writing systems: scripts with clearly separated characters are generally more stable, while ligature-rich or highly connected scripts such as Arabic remain harder to synthesize naturally. Third, diagram generation is currently limited to free-body diagrams and simple sketches rather than complex technical figures. Because the system can imitate a writer from limited samples, it may also enable impersonation, forgery, or misuse in signatures and other identity-sensitive documents. Practical deployment should therefore require appropriate consent and provenance mechanisms and avoid authentication-sensitive use. Future work will expand custom online handwriting data across scripts and tasks, and investigate reinforcement learning for stroke planning, script-specific writing behaviour, and more accurate diagram generation.

\FloatBarrier
\section{Conclusion}
In this paper, we introduced HandwritingAgent, a language-driven handwriting synthesis agent that can discretely generate diverse handwriting styles by reasoning over the geometric structure of written forms through natural language. HandwritingAgent supports both conversational and non-conversational modes of operation, accepting writing styles in image, fonts and stroke formats, while synthesizing discretely in SVG space. By formulating the task as a reasoning-guided symbolic generation problem, the agent can plan glyph formation, adapt stroke geometry, and preserve stylistic visual characteristics without requiring task-specific training. Experimental results show competitive performance against state-of-the-art generative models, especially in visual structure preservation, style imitation, and controllable synthesis across scripts. This highlights the advantage of agentic discrete handwriting synthesis over conventional approaches, with promising applications in education and creative domains.

\bibliography{handwriting-agent-arxiv}
\bibliographystyle{acl_natbib}

\end{document}